\documentclass[conference]{IEEEtran}
\IEEEoverridecommandlockouts
\usepackage{cite}
\usepackage{amsmath,amssymb,amsfonts}
\usepackage{algorithmic}
\usepackage{graphicx}
\usepackage{textcomp}
\usepackage{xcolor}
\def\BibTeX{{\rm B\kern-.05em{\sc i\kern-.025em b}\kern-.08em
    T\kern-.1667em\lower.7ex\hbox{E}\kern-.125emX}}
\begin{document}

\title{How Long short-term memory artificial neural network, synthetic data, and fine-tuning improve the classification of raw EEG data}

\author{
		\IEEEauthorblockN{Albert Nasybullin}
		\IEEEauthorblockA{\textit{Laboratory of Data Analysis}\\
		    \textit{and Bioinformatics,}\\
			\textit{Center for Artificial Intelligence,}\\
			\textit{Innopolis University}\\
			Innopolis,	Russia\\
			a.nasibullin@innopolis.university}
		\and
		\IEEEauthorblockN{Vladimir Maksimenko}
		\IEEEauthorblockA{\textit{Neuroscience and Cognitive}\\ 
			\textit{Technology Laboratory,}\\
			\textit{Center for Technologies in Robotics}\\
			\textit{and Mechatronics Components,}\\
			\textit{Innopolis University}\\
			Innopolis,	Russia\\
			maximenkovl@gmail.com}
	    \and
        \IEEEauthorblockN{Semen Kurkin}
		\IEEEauthorblockA{\textit{Neuroscience and Cognitive}\\ 
			\textit{Technology Laboratory,}\\
			\textit{Center for Technologies in Robotics}\\
			\textit{and Mechatronics Components,}\\
			\textit{Innopolis University}\\
			Innopolis,	Russia\\
			kurkinsa@gmail.com}
		}

\maketitle

\begin{abstract}
In this paper, we discuss a Machine Learning pipeline for the classification of EEG data. We propose a combination of synthetic data generation, long short-term memory artificial neural network (LSTM), and fine-tuning to solve classification problems for experiments with implicit visual stimuli, such as the Necker cube with different levels of ambiguity. The developed approach increased the quality of the classification model of raw EEG data.
\end{abstract}


\section{Introduction}
The brain signal classification is one of the major problems in neurosciences. Current studies propose various methods for EEG data classification. Classical machine learning approaches, such as Support Vector Machine (SVM), Principal Component Analysis (PCA), and k-Nearest Neighbors (k-NN) classifiers, are widely used. Approaches based on artificial neural networks show satisfying results as well\cite{b3, b4, b5}. Successful solving of classification problems is required for extending the development of Brain-Computer Interfaces\cite{b6, b7}.

Age, awareness, caffeine and alcohol consumption, smoking, handedness, and other factors may influence human brain activities\cite{b8,b9}. Due to the generalization ability of artificial neural networks, Deep Learning is a promising direction of study in Brain-Computer Interfaces (BCI). EEG data in neurosciences often suffer from a limited size of experiments. The average number of subjects in neurophysiological experiments is thirty-three participants\cite{b1} and Deep Learning approaches require large size datasets to achieve satisfying results.

In this study, we work with EEG data, using band-pass filtering to generate synthetic datasets, implement LSTM artificial neural networks and use fine-tuning techniques to solve classification problem for the experiment based on visual stimuli and increase classification quality.

\section{Data Collection and Data Structure}

The experimental group contained twenty healthy volunteers of age 26--35, nine participants were female, and eleven participants were male. Participants had a normal or corrected-to-normal vision and participated in the experiment after providing written informed consent. Subjects participated in similar experiments not earlier than six months before. All experiments were carried out under the requirements of the Declaration of Helsinki and approved by the local Research Ethics Committee of the Innopolis University.

\subsection{Visual Stimuli}

\begin{figure*}[htbp]
\centerline{\includegraphics{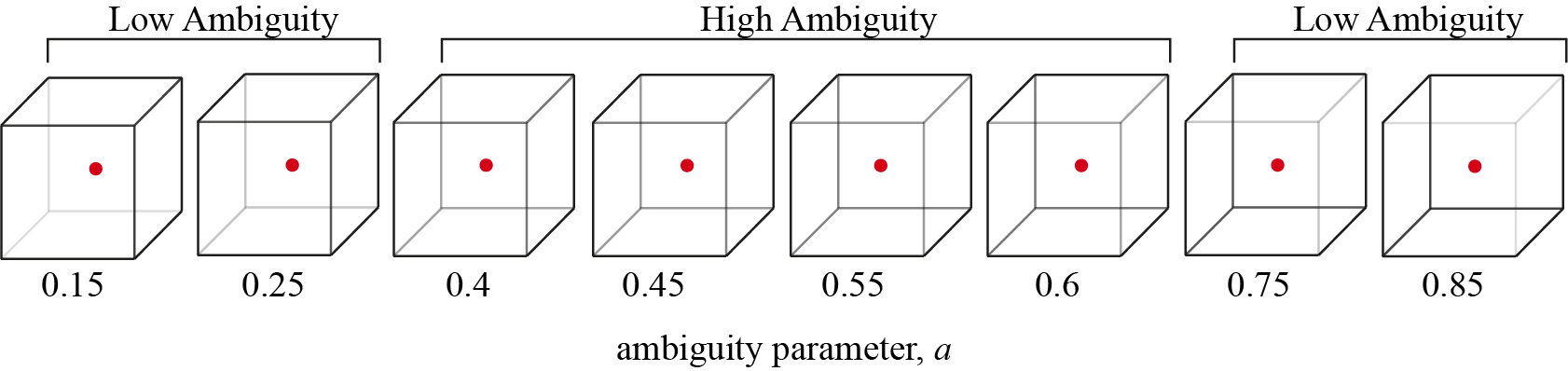}}
\caption{Ambiguous values of the Necker cube in visual stimulus \cite{b2}}
\label{Necker-cubes}
\end{figure*}

In the experiment, the Necker cube served as an ambiguous bistable visual stimulus that may be interpreted in one of two ways. Participants interpreted two-dimensional (2D) images of the Necker cube as three-dimensional (3D) objects with either left or right orientation. The brightness balance between three inner lines of the left-bottom corner and three inner lines in the right upper corner dictates the orientation of the 3D cube and the level of ambiguity. An unambiguous 2D cube is a cube with limiting cases of a = 0 and a = 1 corresponding to the left or the right-oriented cubes, respectively. Completely ambiguous orientation of the 3D defined with a = 0.5. The Necker cube images presented to participants contained set of ambiguity values a = [0.15, 0.25, 0.4, 0.45, 0.55, 0.6, 0.75, 0.85]. Subsets of a = [0.15, 0.25, 0.4, 0.45] and a = [0.55, 0.6, 0.75, 0.85] split set of images to left and right oriented. High ambiguity images are subset of a = [0.40, 0.45, 0.55, 0.60]. The rest of the ambiguous values are considered easily interpretable. Full set of images described in Fig.~\ref{Necker-cubes}.

\subsection{Task}

Images of Necker cubes (22.55 $\times$ 22.55 cm) were shown to participants on a white background on a monitor (52.1 $\times$ 29.3 cm). The resolution of the monitor is 1920 by 1080 pixels and the refresh rate is 60 Hz. Distance from the monitor to a participant is 0.79-0.8 m, visual angle is {$\sim$}0.39 rad. The duration of the experiment for each participant is about 40 min. Before and after the main part of the experiment eyes-open resting state was recorded for $\approx$150 s. Subjects looked at about 500 Necker cube images with predefined levels of ambiguity in random order. Each image appeared on the screen for 1-1.5 s. White noise pictures appeared on the screen for 3-5 s between Necker cube images.
The participants were instructed to respond to the left or right stimulus orientation of the visual stimulus by pressing the left or right key, respectively. We monitored presentation time (when a stimulus appeared) and decision time (time passed from stimulus appearance to key pressing). To extract participant decisions' error rates, we matched participants' key responses and actual ambiguity values. Subjects with an error rate higher, than 20\% were excluded from the dataset. 

\subsection{EEG data acquisition}

We used a classical 10--10 electrode scheme to register the EEG signal. We recorded signals from thirty-one channels using an electrode cap, with two reference electrodes on the earlobes (A1 and A2) and a ground electrode N above the forehead. Ag/AgCl cup adhesive electrodes placed on the ``Tien--20'' paste (Weaver and Company, Aurora, CO, USA) were used for signal acquisition. Immediately before the experiments, a special abrasive ``NuPrep'' gel (Weaver and Company, Aurora CO, USA) was applied to the electrode attachment areas to increase skin conductivity. We maintained the impedance values in the range of 2--5 k$\Omega$. For registration, amplification, and analog-to-digital conversion of the EEG signals, we used a multichannel electroencephalograph ``Encephalan-EEG-19/26'' (Medicom MTD company, Taganrog, Russian Federation) with a two-button input device (keypad). This device holds the registration certificate from the Federal Service for Supervision in Health Care No. FCP 2007/00124 of 07.11.2014 and European Certificate CE 538571 from the British Standards Institute (BSI).

\subsection{Data Structure}

The experimental dataset contains 4 000 files in .CSV format with a split to each level of ambiguity mentioned above a = [0.15, 0.25, 0.4, 0.45, 0.55, 0.6, 0.75, 0.85]. Each file contains thirty-one rows and from 700 to 1500 columns, due to the variable length of trials in the experiment. Each row corresponds to a different channel. In this study, we separate the dataset into three classes with correspondence to different levels of ambiguity. Subsets a = [0.15, 0.25] and a = [0.75, 0.85] define left-oriented (997 trials) and right-oriented (1 003 trials) cube classes, respectively. Subset a = [0.4, 0.45, 0.55, 0.6] defines the high-ambiguity (2 000 trials) class. These conditions make our work a multi-class classification problem of imbalanced data.

Dataset was split into 60/40 proportions. The conventional train/test split in Deep Learning is 80/20\cite{b10}. We chose it for the demonstration purposes of the proposed method. To have an equal distribution of participants in the test set, we used K-fold Cross-Validation with twenty folds. Table~\ref{Original dataset structure} shows structure of original dataset.

\begin{table}[htbp]
\caption{Original dataset structure}
\begin{center}
\begin{tabular}{|c|c|c|c|}
\hline
\textbf{Class labels} & "Left" & "High-ambiguity" & "Right" \\ \hline
\textbf{Ambiguity values} & 0.15, 0.25 & 0.4, 0.45, 0.55, 0.6 & 0.75, 0.85 \\ \hline
\textbf{Total dataset size} & 997 & 2 000 & 1 003 \\ \hline
\textbf{Train subset size} & 598 & 1 200 & 602 \\ \hline
\textbf{Test subset size} & 399 & 800 & 401 \\ \hline
\end{tabular}
\label{Original dataset structure}
\end{center}
\end{table}

\section{LSTM neural network, synthetic data generation, and fine-tuning}

\subsection{LSTM neural network model}

The long short-term memory model of the artificial neural network was first introduced in 1997\cite{b11}. LSTM is a type of recurrent neural network and can process not only single data points but sequences of data. LSTM models are used for handwriting recognition, speech recognition, robot control, video prediction, and neuroscience problems as well\cite{lstm1, lstm2, lstm3, lstm4}. We developed the model architecture and parameters empirically during the study.

\begin{figure*}[htbp]
\centerline{\includegraphics[width=1\textwidth]{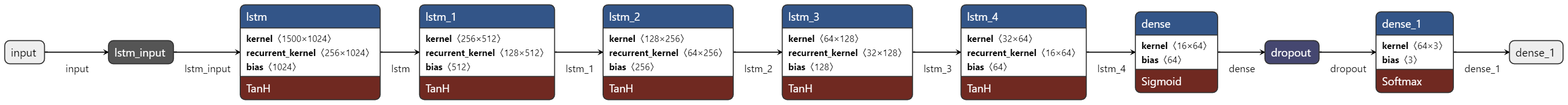}}
\caption{The architecture of the Long short-term memory artificial neural network model used for classification.}
\label{model}
\end{figure*}

The input shape of the LSTM model was thirty-one by 256 (see Fig.~\ref{model}). The model contains five LSTM layers with decreasing densities of 256, 128, 64, 32, and sixteen artificial neurons, respectively. LSTM layers one-four return sequences, the LSTM layer five does not return sequences. Layer six is a fully connected layer of sixty-four artificial neurons and sigmoid activation function. Layer number seven is a fully connected layer with three artificial neurons corresponding to the number of classes and SoftMax activation function. We apply a Dropout of rate 0.3 between layers six and seven. Sparce categorical cross-entropy is a loss function. The total number of parameters in the model is 2 062 531.

\subsection{LSTM classification of original data}

To introduce our method, we apply the developed model to the original data and will compare the results between model performances. Classes on the proposed dataset differ in the number of trials. We can consider our problem imbalanced data multi-class classification problem. To evaluate the model's performance, we used a weighted-average F1 score\cite{f1}. All demonstrated results are taken from the first attempt of the experiment. We did not use so-called ``Cherry-Picking'' in this work\cite{cherry}.

We trained the LSTM model for 100 epochs and reached a weighted-average F1 score of 0.63. Taking into account the complexity of the problem and the type of stimulus presented to participants, we can consider those results satisfying.

\subsection{Expanding the size of the dataset with band-pass filtering and pre-training LSTM model}

To improve the results of classification and avoid overfitting issues, we propose to apply fine-tuning method with synthetic data. Fine-tuning implies the use of a trained artificial neural network and the use of its weights as initial weights for a new model of the same domain. Fine-tuning implies overcoming a small dataset size problem.

In our study, we use a band-pass filter to create a new dataset and pre-train our model. The size of the original dataset is 4 000 trials. We applied three different band-pass filters to create a dataset of 12 000 trials. Each filter represents a frequency range that usually corresponds to a different type of brain activity and creates an additional data subset of 4 000 trials: theta frequency (4--8 Hz), alpha frequency (8--14 Hz), and beta frequency (14--30 Hz). The generated dataset has a size of 12 000 trials: the left-oriented cube's class is 3 988 trials, the right-oriented cube's class is 4 012 trials and the high-ambiguity class is 8 000 trials.

\begin{figure}[htbp]
\centerline{\includegraphics[width=0.5\textwidth]{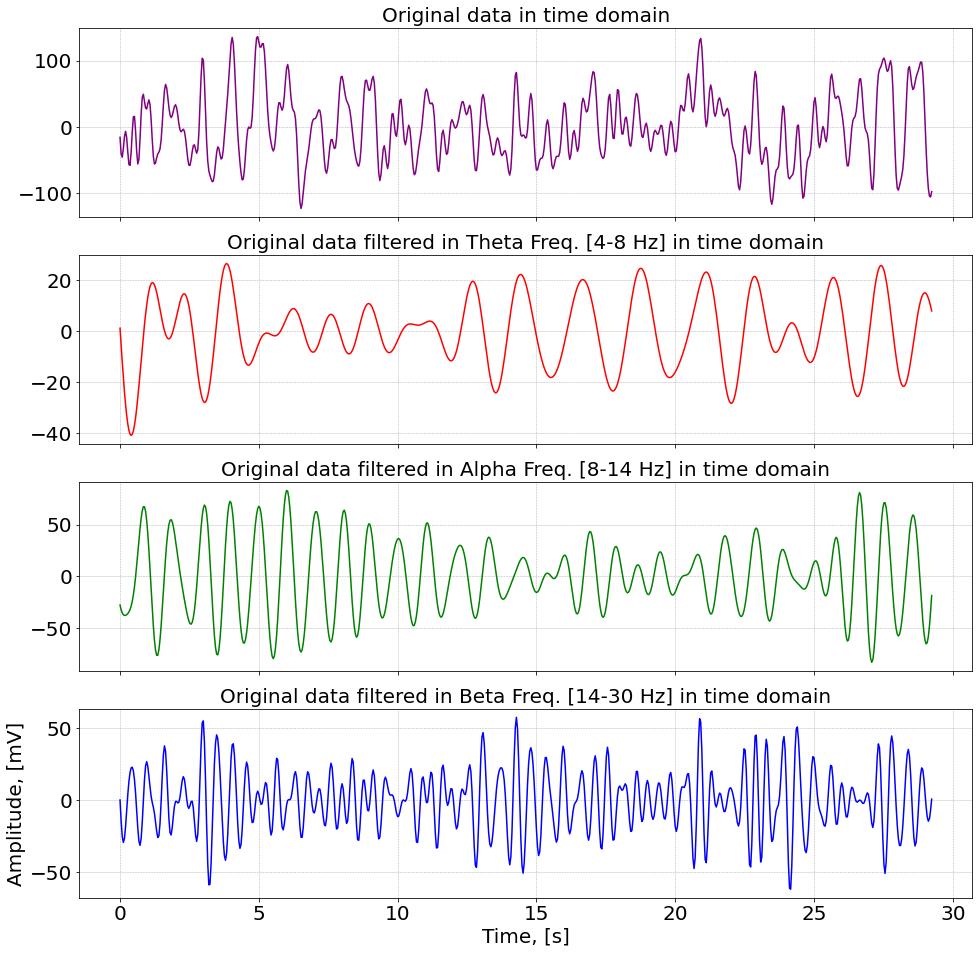}}
\caption{We build our intuition behind data expanding strategy on a difference created in a signal by band-pass filtering.}
\label{combined}
\end{figure}

We trained the LSTM model for twenty epochs for each generated subset. First, we pre-trained the model on the theta frequency subset and saved pre-trained weights. Then, we initialize the new model with pre-trained weights and continue training with the alpha frequency subset. The same logic appears in each created subset. In the end, we got weights of pre-trained models on different combinations of subsets.

\subsection{LSTM classification with pre-trained models}

We have trained the same model with different initial weights on fifty epochs each time. The model achieved the highest value of the weighted-average F1 score with initial weights from the theta frequency subset (F1 score = 0.78). Weighted-average F1 scores for the rest of the pre-trained weights are lower but surpassed the results from the original data. All experimental results described in Table~\ref{Model evaluation with different pre-trained weights}.

The higher performance of the model pre-trained on theta subset is hard to interpret and requires additional study.

\begin{table}[htbp]
\caption{Model evaluation with different pre-trained weights}
\begin{center}
\begin{tabular}{|c|c|c|c|c|}
\hline
\textbf{}&\multicolumn{4}{|c|}{\textbf{Training data subset}} \\ \hline
\cline{2-5}
\textbf{F-1 score} & Original & Theta & Alpha & Beta \\ \hline
\textbf{0.63} & + &  &  &  \\ \hline
\textbf{0.78} & + & + &  &  \\ \hline
\textbf{0.71} & + &  & + &  \\ \hline
\textbf{0.69} & + &  &  & + \\ \hline
\textbf{0.73} & + & + & + & + \\ \hline
\end{tabular}
\label{Model evaluation with different pre-trained weights}
\end{center}
\end{table}

\section{Discussion}

The proposed method extends the classification capabilities of raw EEG data in the case of ambiguous stimuli with artificial deep neural networks. We expect the proposed method to be relevant for different architectures of artificial neural networks. We will continue our study of raw EEG data classification.

Future work shall answer several questions. Can the method be used for eight classes of ambiguity defined by the original data? Can we use the method for the feature extraction problem? Why does a model pre-trained on theta subset achieve better results?

\section{Conclusion}

In the work, we explored the combination of data expansion with band-pass filtering and fine-tuning for the EEG data classification problem. Despite the high ambiguity of the stimuli presented to participants, the proposed method increases the quality of the classification model. All pre-trained models increase classification ability. However, the model pre-trained on the theta frequency subset reaches the highest classification results.

\section*{Acknowledgment}

This work has been supported by the Center for Technologies in Robotics and Mechatronics Components (Innopolis University). A.N. acknowledges The Analytical Center for the Government of the Russian Federation (Agreement No. 70-2021-00143 dd. 01.11.2021, IGK 000000D730321P5Q0002) for the support.

\vspace{12pt}


\begin{thebibliography}{00}

\bibitem{b3}P. Chholak, A. N. Pisarchik, S. A. Kurkin, V. A. Maksimenko, and A. E. Hramov, ``Phase-amplitude coupling between mu- and gamma-waves to carry motor commands,'' in 2019 3rd School on Dynamics of Complex Networks and their Application in Intellectual Robotics (DCNAIR), 2019, pp. 39--45.

\bibitem{b4}S. Kurkin, P. Chholak, V. Maksimenko, and A. Pisarchik, ``Machine learning approaches for classification of imaginary movement type by MEG data for neurorehabilitation,'' in 2019 3rd School on Dynamics of Complex Networks and their Application in Intellectual Robotics (DCNAIR), 2019, pp. 106--108.

\bibitem{b5}S. Y. Gordleeva et al., ``Real-time EEG--EMG human--machine interface-based control system for a lower-limb exoskeleton,'' IEEE Access, vol. 8, pp. 84070--84081, 2020.

\bibitem{b6}S. P. Liburkina, A. N. Vasilyev, L. V. Yakovlev, S. Y. Gordleeva, and A. Y. Kaplan, ``A motor imagery-based brain--computer interface with vibrotactile stimuli,'' Neurosci. Behav. Physiol., vol. 48, no. 9, pp. 1067--1077, 2018.

\bibitem{b7}U. Chaudhary et al., ``Spelling interface using intracortical signals in a completely locked-in patient enabled via auditory neurofeedback training,'' Nat. Commun., vol. 13, no. 1, p. 1236, 2022.

\bibitem{b8}N. S. Frolov et al., ``Age-related slowing down in the motor initiation in elderly adults,'' PLoS One, vol. 15, no. 9, p. e0233942, 2020.

\bibitem{b9}B. Fiani et al., ``The neurophysiology of caffeine as a central nervous system stimulant and the resultant effects on cognitive function,'' Cureus, vol. 13, no. 5, p. e15032, 2021.

\bibitem{b1}A. Bazzani, S. Ravaioli, L. Trieste, U. Faraguna, and G. Turchetti, ``Is EEG suitable for marketing research? A systematic review,'' Front. Neurosci., vol. 14, p. 594566, 2020.

\bibitem{b2}A. K. Kuc, S. A. Kurkin, V. A. Maksimenko, A. N. Pisarchik, and A. E. Hramov, ``Monitoring brain state and behavioral performance during repetitive visual stimulation,'' Appl. Sci. (Basel), vol. 11, no. 23, p. 11544, 2021.

\bibitem{b10}K. M. Kahloot and P. Ekler, ``Algorithmic splitting: A method for dataset preparation,'' IEEE Access, vol. 9, pp. 125229--125237, 2021.

\bibitem{b11}S. Hochreiter and J. Schmidhuber, ``Long short-term memory,'' Neural Comput., vol. 9, no. 8, pp. 1735--1780, 1997.

\bibitem{lstm1}A. Graves, M. Liwicki, S. Fern\'andez, R. Bertolami, H. Bunke, and J. Schmidhuber, ``A novel connectionist system for unconstrained handwriting recognition,'' IEEE Trans. Pattern Anal. Mach. Intell., vol. 31, no. 5, pp. 855--868, 2009.

\bibitem{lstm2}H. Sak, A. Senior, and F. Beaufays, ``Long short-term memory recurrent neural network architectures for large scale acoustic modeling,'' in Interspeech 2014, 2014.

\bibitem{lstm3}M. J. Monesi, B. Accou, J. Montoya-Martinez, T. Francart, and H. Van Hamme, ``An LSTM based architecture to relate speech stimulus to EEG,'' in ICASSP 2020 - 2020 IEEE International Conference on Acoustics, Speech and Signal Processing (ICASSP), 2020, pp. 941--945.

\bibitem{lstm4}N. Elsayed, A. S. Maida, and M. Bayoumi, ``Reduced-gate convolutional LSTM architecture for next-frame video prediction using predictive coding,'' in 2019 International Joint Conference on Neural Networks (IJCNN), 2019, pp. 1--9.

\bibitem{f1}A. Tharwat, ``Classification assessment methods,'' Appl. Comput. Inform., vol. 17, no. 1, pp. 168--192, 2021.

\bibitem{cherry}J. Komiyama and T. Maehara, ``A simple way to deal with cherry-picking,'' arXiv [stat.ME], 2018.

\end{thebibliography}
\end{document}